\documentclass[conference]{IEEEtran}
\usepackage{cite}
\usepackage{amsmath,amssymb,amsfonts}
\usepackage{algorithmic}
\usepackage{graphicx}
\usepackage{textcomp}
\usepackage{hyperref}
\def\BibTeX{{\rm B\kern-.05em{\sc i\kern-.025em b}\kern-.08em
    T\kern-.1667em\lower.7ex\hbox{E}\kern-.125emX}}
\begin{document}

\title{Improved Text Language Identification for the South African Languages\\
}


\author{
\IEEEauthorblockN{Bernardt Duvenhage}
\IEEEauthorblockA{bernardt@feersum.io \\
\textit{Feersum Engine} \\
\textit{Praekelt Consulting}\\
Johannesburg, South Africa}
\and
\IEEEauthorblockN{Mfundo Ntini}
\IEEEauthorblockA{mfundo@praekelt.com \\
\textit{Engineering Team} \\
\textit{Praekelt Consulting}\\
Johannesburg, South Africa}
\and
\IEEEauthorblockN{Phala Ramonyai}
\IEEEauthorblockA{phala@praekelt.com \\
\textit{Engineering Team} \\
\textit{Praekelt Consulting}\\
Johannesburg, South Africa}

}

\maketitle

\begin{abstract}
Virtual assistants and text chatbots have recently been gaining popularity. Given the short message nature of text-based chat interactions, the language identification systems of these bots might only have 15 or 20 characters to make a prediction. However, accurate text language identification is important, especially in the early stages of many multilingual natural language processing pipelines.

This paper investigates the use of a naive Bayes classifier, to accurately predict the language family that a piece of text belongs to, combined with a lexicon based classifier to distinguish the specific South African language that the text is written in. This approach leads to a 31\% reduction in the language detection error.

In the spirit of reproducible research the training and testing datasets as well as the code are published on github. Hopefully it will be useful to create a text language identification shared task for South African languages.
\end{abstract}


\section{Introduction}
\label{sec:intro}
Virtual assistants and text chatbots seem to be gaining much popularity, but to be accessible to South Africans these software agents need to understand our local languages. South Africa has 11 official languages belonging to a couple different language families. Afrikaans (afr) and English (eng) are Germanic languages. isiNdebele (nbl),  isiXhosa (xho), isiZulu (zul) and siSwati (ssw) belong to the Nguni family of languages. Sepedi (nso), Sesotho (sot) and Setswana (tsn) belong to the Sotho-Tswana family of languages. Finally, Xitsonga (tso) belong to the Tswa-Ronga family and Tshivenda (ven) belong to the Venda family. Many of these languages are under-resourced and further work is required to build software agents that are fluent in the country's rich vernacular.

Text language identification (LID) is an important early step in many multi-lingual natural language processing (NLP) pipelines because many of the later steps are still language dependent. Given the short message nature of text based chat interactions and the possibility of code switching the language identification system might only have 15 or 20 characters to make a prediction. However, lower LID accuracies may be expected for short text due to fewer text features being available during classification. Any errors that occur early in an NLP pipeline are also potentially compounded by later processing steps.

This paper gives an overview of the related LID literature in Section~\ref{sec:related_works}, a discussion of the chosen baseline classifier in Section~\ref{sec:baseline} followed by the discussion of the paper's contribution to the improvement of LID on short pieces of text in Section~\ref{sec:core}. Comparative results are presented in Section~\ref{sec:results} followed by some concluding remarks and suggested future work in Section~\ref{sec:conclusion}.

A further contribution of this work is that the training and testing datasets as well as the code are published on github. Hopefully it will be useful to create a text language identification shared task for South African languages.

\begin{figure*}[!t]
\centering
\includegraphics[width=2.0\columnwidth]{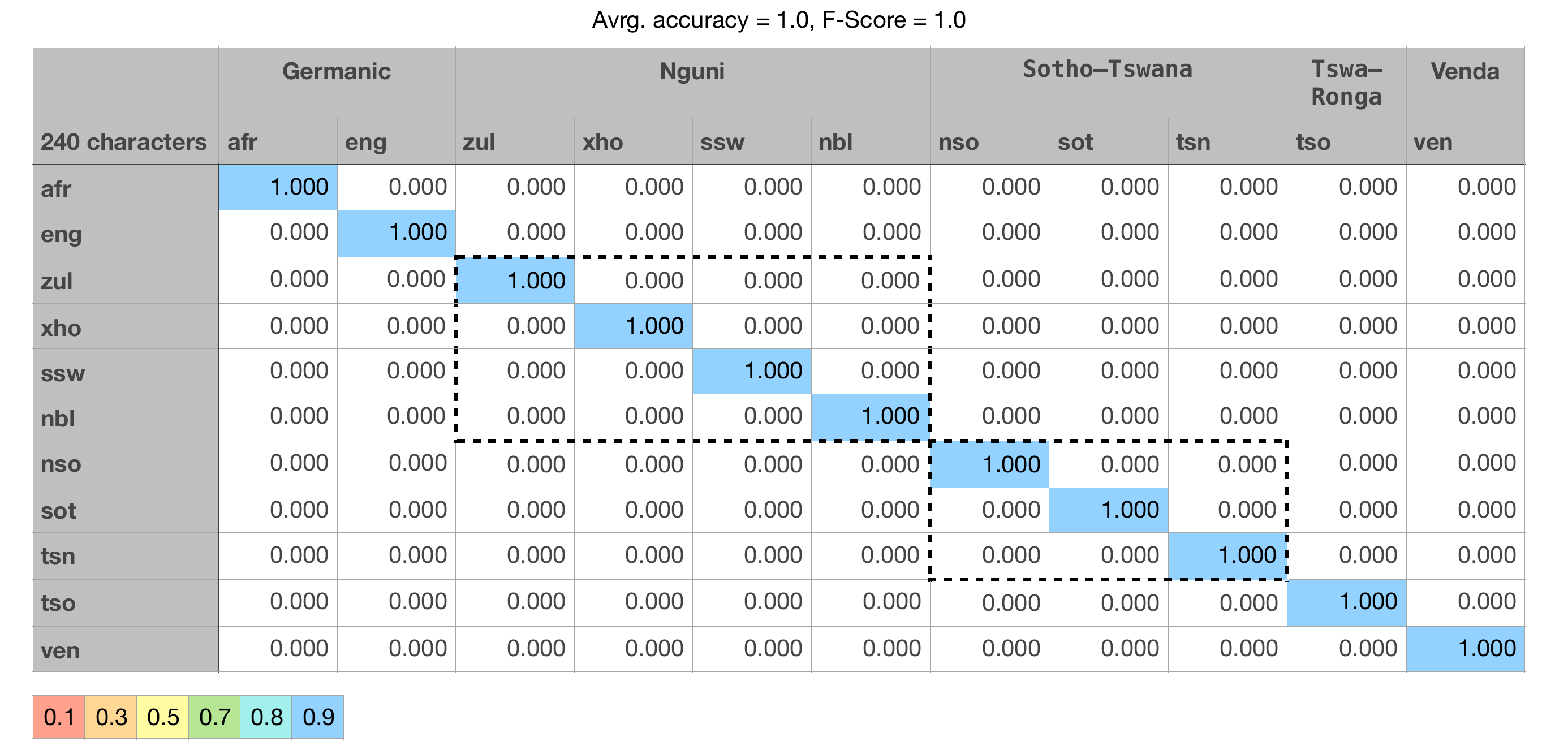}
\caption{Confusion matrix of the baseline classifier and a test set with strings of length 200-300 characters.}
\label{fig:cm_clean_300}
\end{figure*}


\section{Related works}
\label{sec:related_works}
An LID system for long texts based on normalised histograms of character n-grams is presented in \cite{CombrinckBotha1995}. A similar system that also successfully used character n-grams for doing LID of long texts is presented in \cite{CavnerTrenkle1997}

A frequency based n-gram difference based classifier and a support vector machine (SVM) that uses the n-gram frequencies as features are discussed in \cite{BothaZimuBarnard2006}. Error rates of approximately 0.3\% are achieved over large text window sizes. It is also found that the SVM's performance is better than the n-gram based estimator's, but at a much greater computational cost.

In \cite{PienaarSnyman2010} a spell-checker from the South African Centre for Text Technology (CTexT) is applied to do LID. A sentence level accuracy of 97.9\% is achieved on texts of approximately 400 characters in length.

A naive Bayes classifier with various character n-gram text features, called \emph{langid}, is discussed in \cite{LuiBaldwin2012}. In the current paper langid is also trained on the South African languages and used as an LID reference in Section~\ref{sec:results}.

A difference in n-gram frequencies classifier, a naive Bayes text classifier with n-gram features and an SVM are evaluated for LID in \cite{BothaBarnard2012}. The Bayesian classifier is reported to be the most accurate in practise at 17\% error on texts of 15 characters. 

An SVM and a naive Bayes classifier for language identification of individual words are compared in \cite{GiwaDavel2013}. The system was trained to identify afr, eng, sot and zul which, except for afr and eng, are all from different South African language families. Accuracies of around 85\% - 95\% on single 10 - 15 character words are reported.


In \cite{GiwaDavel2014} joint sequence models are used to further improve the accuracy of LID of single words 10 - 15 characters long. Accuracies of around 97.2\% are reported when labelling text as afr, eng, sot or zul. The training data used is from the National Centre for Human language Technology's language dictionary word lists.

In \cite{JaechEtAl2016} Char2Vec and an LSTM are used to do end-to-end trained LID. Char2Vec is used to get word embeddings which are then combined via an LSTM. Once trained the LSTM is able to predict a language for each word in the sentence. The significant text features are automatically learned and text pre-processing and cleanup is not required. Near state-of-the-art performance is reported on code switching LID shared tasks.

Recently a lexicon based LID~\cite{SelamatAkosu2016} was applied to under-resourced languages. It is not clear what the character length of the training and testing samples were, but the reported LID accuracies are in the low 90's.


\begin{figure*}[!t]
\centering
\includegraphics[width=2.0\columnwidth]{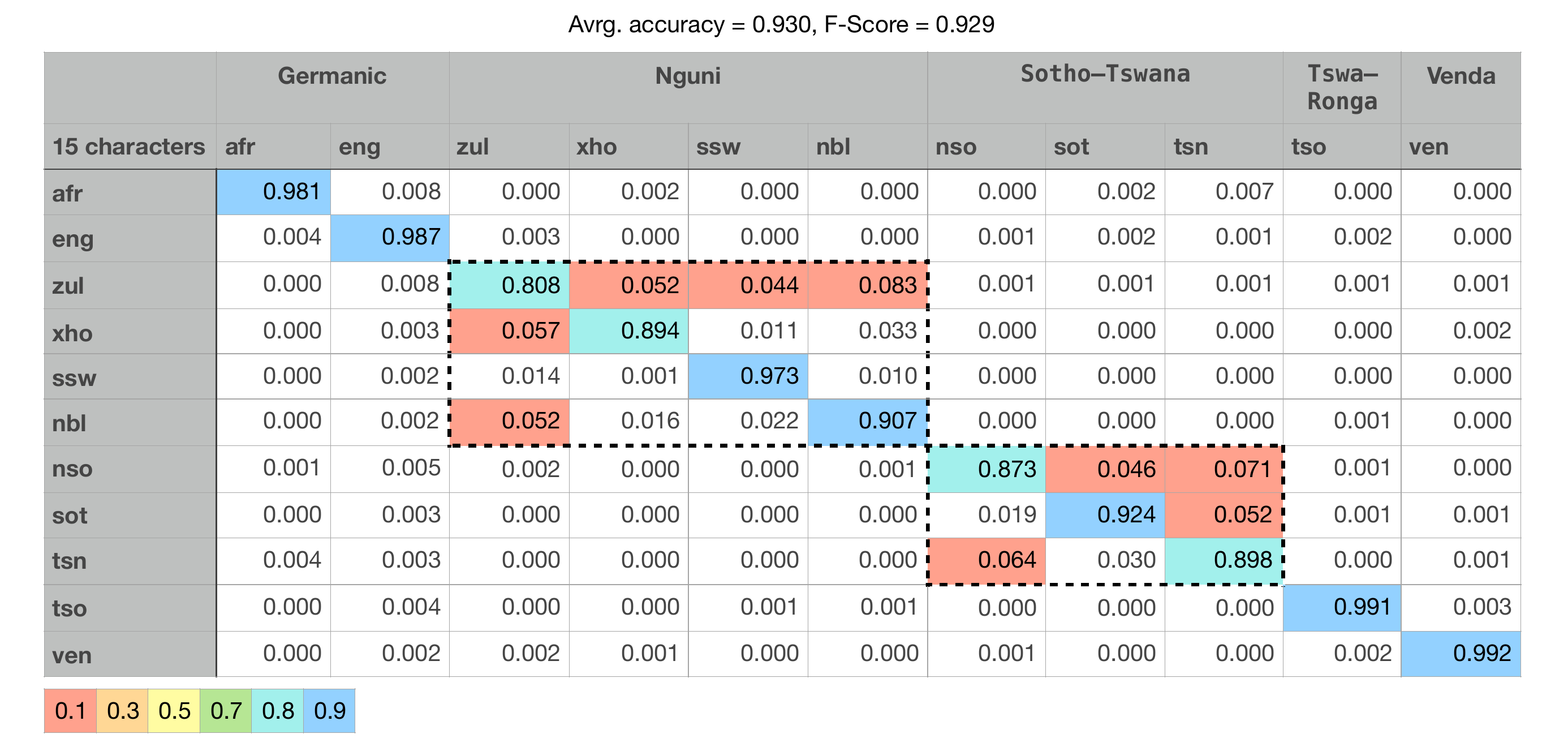}
\caption{Confusion matrix of the baseline classifier and a test set with strings of length 15 characters.}
\label{fig:cm_clean_15}
\end{figure*}

\section{LID Baseline for South African Languages Using Character N-Grams}
\label{sec:baseline}
The use of a naive Bayes classifier with character n-gram text features has become the standard for sentence level text LID~\cite{BothaBarnard2012}. The baseline used in this paper is sklearn's multinomial naive Bayes classifier with character 5-grams. 

The data used in this paper is the NCHLT Text Corpora~\cite{NCHLTTextCorpora}\cite{EiselenPuttkammer2014}. The NCHLT Text Corpora data was cleaned up a bit by replacing numbers and all punctuation except '-' with spaces. All other characters such as \v{s} were left unmodified.

\begin{figure}[!t]
\centering
\includegraphics[width=1.0\columnwidth]{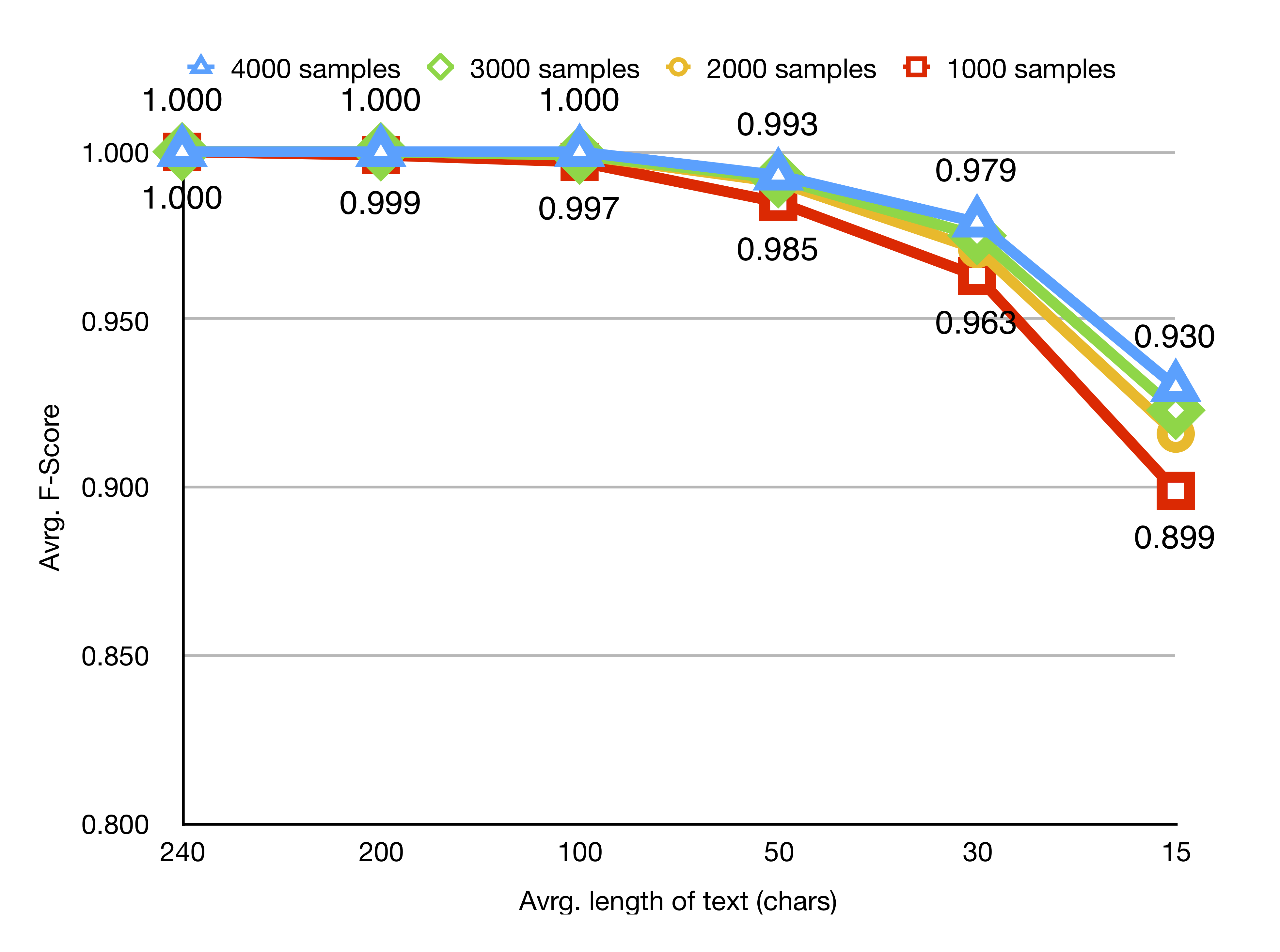}
\caption{The baseline classifier's F-score for shorter text fragments. The different graphs show how the number of full sentence training samples influence the short sentence LID results.}
\label{fig:f_score_clean}
\end{figure}

3000 training samples and 1000 test samples per language were randomly chosen from the subset of full sentences in the CText data that are 200-300 characters long. Using more training data doesn't significantly improve the LID accuracy for long sentences as shown later in Figure~\ref{fig:f_score_clean} in the results section. Binary text features were used as opposed to integer feature counts. Later in the paper a classifier trained on all 4000 of these long sentences are reused for classification of short sentences.

Initially the trained classifier had an accuracy and F-score of 99.5\%. However, some of the mis-predicted sentences were spotted to be mislabelled in the NCHLT data. The mis-predicted sentences were few enough to all be checked manually and were indeed found to be mislabelled in the CTexT data. Approximately 0.468\% of the data was correctly relabeled in this manner. The updated datasets are hosted with the LID code on github at \url{https://github.com/praekelt/feersum-lid-shared-task}.

After cleaning up the data the trained classifier had an accuracy and F-score of 99.9909\% $\approx$ 100.0\%. This baseline classifier already outperforms previous work~\cite{BothaBarnard2012} on long sentences.~\footnote{The authors of the earlier work didn't attribute their data to CTEXT, but it is curious that they achieved the same accuracy as we did before the data cleanup.} Figure~\ref{fig:cm_clean_300} shows the confusion matrix for the test set.

The Google Translate API was also used to verify the results of the n-gram classifier for the languages it understands (i.e. afr, eng, sot, xho and zul). The Google results correlated with all predictions except for some differences between isiXhosa and isiZulu - which belong to the same language family. Approximately 0.09\% of the Google results differed from the baseline results, but again all of these could be checked manually and were found to have been incorrectly labelled by Google Translate. A side effect of this validation is that one can be assured that Google's API is relatively accurate for the South African languages it supports.

The baseline model trains and tests in 90 minutes on a single core of a 3.30 GHz i5 CPU, uses below 2GB of RAM during training and the trained model is approximately 50MB in size. Long sentence language detection therefore seems to be a solved problem for at least the 11 official South African languages and given that the training data is from a similar domain as one's production environment. Others~\cite{GiwaDavel2013} have also noted that one easily achieves 100\% LID accuracy given 300 characters of text.


\section{LID of Short Strings}
\label{sec:core}
Short sentence data was derived from the cleaned dataset by selecting from the start of the long sentences 200, 100, 50, 30 or 15 characters plus any characters required to not split a word. A potential problem with the alternative sliding window approach that others have used is that the fragmented start and end words affect the classifier performance for short sentences and such an approach would also prohibit the use of a lexicon based classification algorithm.

The classifier's F-score on short pieces of text are shown in Figure~\ref{fig:f_score_clean} for training set sizes from 1000 to 4000 samples. The datasets size prevents using more samples to train the baseline classifier, but from the graph it seems that the short sentence performance could benefit from using more than 4000 training samples. Although not the focus of this paper note that our baseline already outperforms earlier reported results~\cite{BothaBarnard2012} of 1.5\% on 100 char strings and 17\% for 15 chars. The current baseline achieves 0.1\% error on 100 char strings and 7.0\% error for 15 chars.


Figure~\ref{fig:cm_clean_15} shows the confusion matrix for short 15 char strings. As others have noted, there is some confusion between languages of the same family. This can be clearly seen from the widening of the diagonal into the family blocks. Note also the limited confusion between language families.

\begin{figure}[!t]
\centering
\includegraphics[width=1.0\columnwidth]{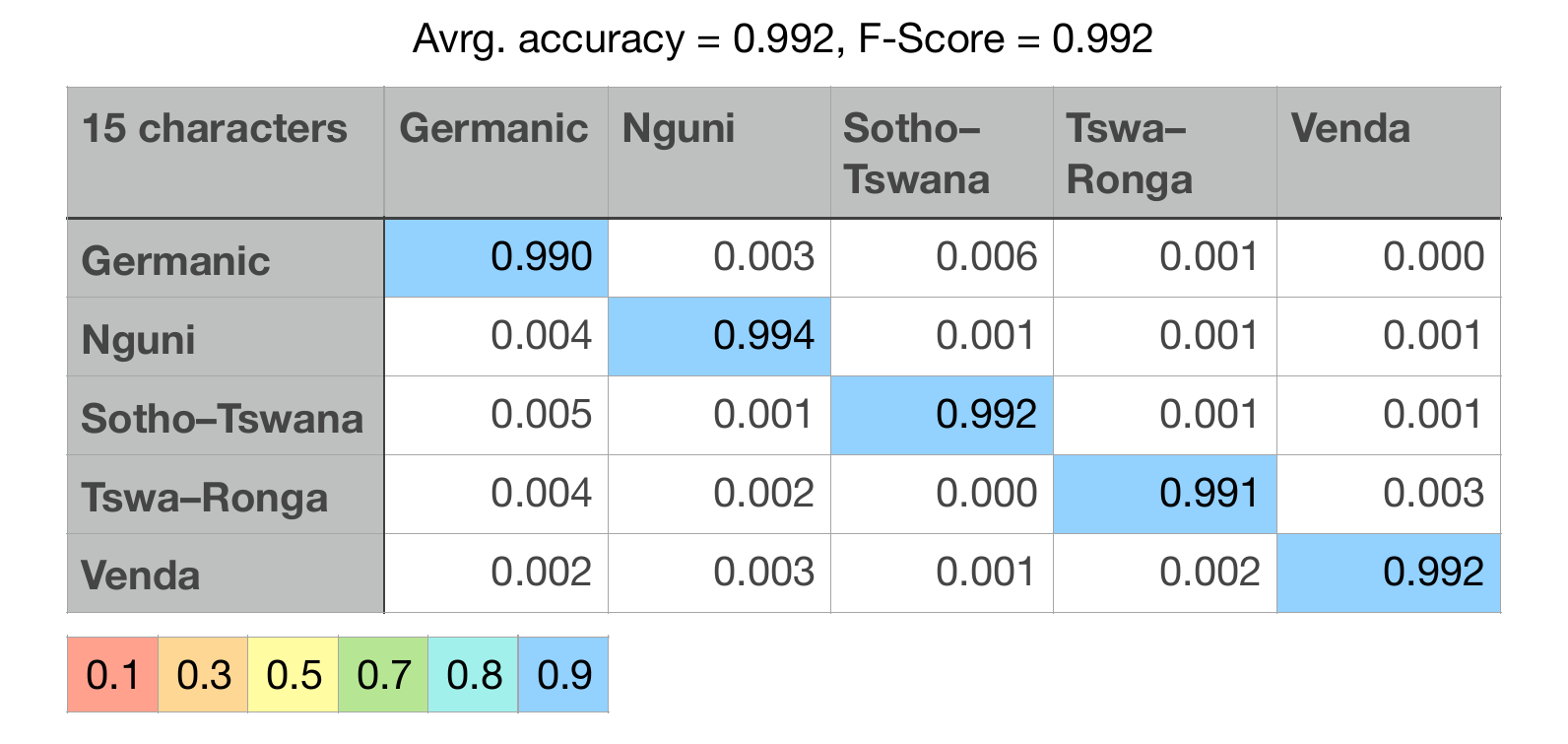}
\caption{Confusion matrix of language families of a test set with strings of length 15 characters.}
\label{fig:cm_clean_15_fam}
\end{figure}

Figure~\ref{fig:cm_clean_15_fam} shows the family confusion matrix for short sentences of 15 characters. The limited confusion between language families should be clear. Also note that the average accuracy and F-score of classifying a short string into language families is 99.2\% while the average accuracy of classifying a short string all the way into a language is only 93.0\%

The baseline classifier therefore performs very well (99.2\% accurate) at classifying even short 15 character sentences into their language families. Such accurate classification of the language family is possibly good enough to enable a software agent to interpret and act on short sentences. However, in some cases one might want to identify the specific language that a piece of text is written in.

A key realisation is that although Sesotho, Setswana and Sepedi are strongly related, certain words might appear in one language, but not in the others within the family. The same is true of isiZulu, siSwati, isiXhosa and isiNdebele. Therefore, a second lexicon based classifier may be useful to distinguish languages within the same language family.

To test this idea, a lexicon is created from all the sentences in the cleaned language corpuses (4.1k - 25k samples per language). During language identification the naive Bayes classifier result is used to classify the text as belonging to a language family after which the language lexicons are used to count how many words of each language in the family is present in the input. If one language in the family dominates then it is chosen as the language label otherwise the naive Bayes result is taken as the most informative and used as the language label.

\section{More Results and Analysis}
\label{sec:results}

\begin{figure}[!t]
\centering
\includegraphics[width=1.0\columnwidth]{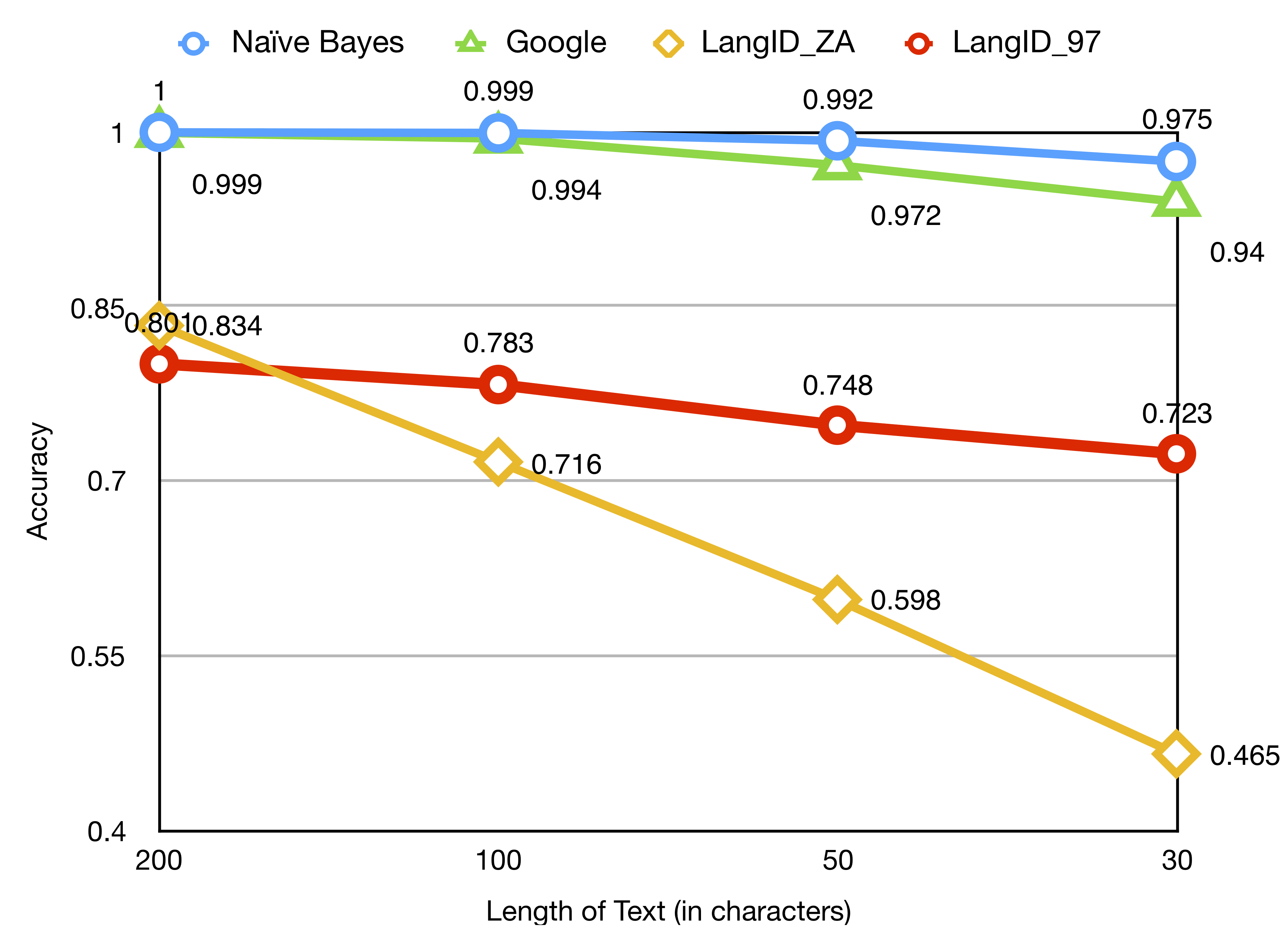}
\caption{Accuracy of the baseline NB, Google Translate, langid\_ZA (trained on cleaned data) and langid\_97 (built-in langid model trained on 97 languages).}
\label{fig:accuracy_compare}
\end{figure}

Figure~\ref{fig:accuracy_compare} shows the comparative accuracies of langid\_97, langid\_za, Google Translate's language detection API and the naive Bayes baseline classifiers. langid\_97 is the langid model included with the langid package that was trained on 97 languages. langid\_97 and Google translate's detector are pretrained and were tested on the full 4000 (training + testing) cleaned samples per language. The South African languages that langid\_97 supports are afr, eng, xho and zul. Google's detector additionally supports sot. For the pre-trained models only the supported languages were included in the accuracy and F-score estimates. The other langid model, langid\_za, we trained on the cleaned-up long (200-300 character) full sentences used in this paper. An ideal accuracy for language identification is above 99\% so that less than one in a hundred predictions fail.

\begin{figure*}[!t]
\centering
\includegraphics[width=2.0\columnwidth]{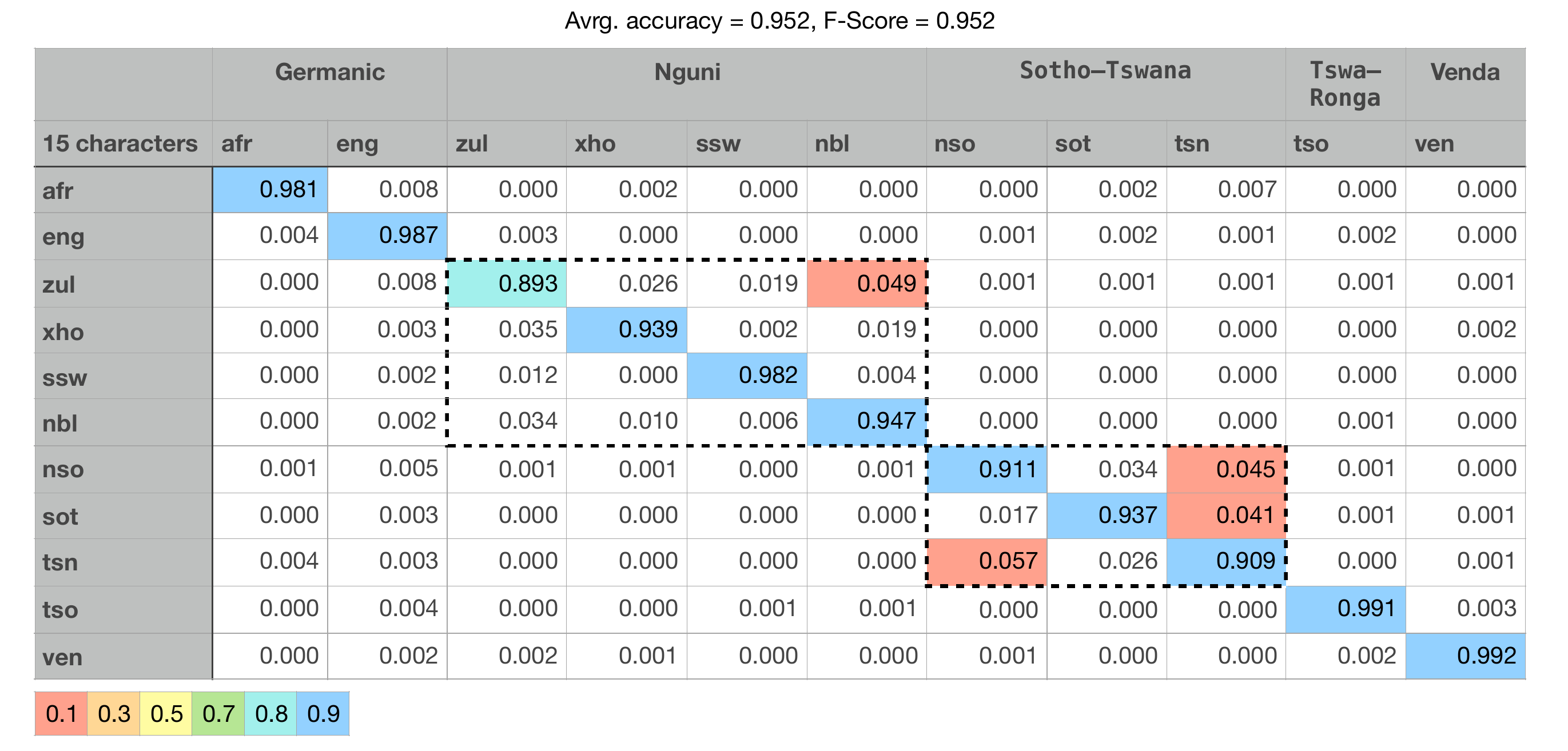}
\caption{Confusion matrix of the lexicon improved classifier and a test set with strings of length 15 characters.}
\label{fig:cm_clean_15_nb_lexi}
\end{figure*}

The proposed lexicon classifier on its own achieves an accuracy of only 89.8\% and an F-score of 89.7\%. However, when used in combination with the baseline classifier the lexicon classifier's result is only used when it responds with a high confidence which results in an overall reduction in error. Figure~\ref{fig:cm_clean_15_nb_lexi} shows the updated confusion matrix for short 15 char strings classified using the simple two stage classifier. The noise level in the results using the 1000 testing samples seem to be in the order of 0.001.

The resulting short sentence LID accuracy is 95.2\% which is 31\% reduction in LID error over the baseline classifier. The family LID accuracy stays unchanged at 99.2\% accuracy. As mentioned previously, earlier works by Botha and Barnard~\cite{BothaBarnard2012} did a full language classification, but only achieved a 17\% error (83\% accuracy) for short sentences of 15 characters. Giwa and Davel~\cite{GiwaDavel2014} achieved what is essentially a language family LID accuracy of just over 97\% for single words of 10 - 15 characters long.


\section{Conclusion}
\label{sec:conclusion}
\subsection{Summary}
The baseline naive Bayes classifier is shown to be more accurate than Google Translate's pre-trained language identification API, the pre-trained langid\_97 and a langid model trained on the cleaned data used in this paper. The baseline also outperforms earlier reported results on South African languages as discussed in Section~\ref{sec:core}.

Adding a lexicon to the baseline classifier reduced LID error by 31\%. The resulting short sentence LID accuracy is 95.2\%. When compared to the previous reported result~\cite{BothaBarnard2012} of 83\% the current model reduced the error from 17\% to 4.8\% which is a 3x reduction in error. 

The improved dataset and code are hosted at \url{https://github.com/praekelt/feersum-lid-shared-task}. It is possible that the process of shortening a sentence changes the certainty of its language label when distinguishing words and other features are are lost. This would result in a performance ceiling for short sentence LID.

\subsection{Future work}
The Multinomial NB classifier was used with binary n-gram features. It should be interesting to compare the results of using the normalised feature counts as well.

The short sentence language labels need to be verified and it is important to also gather data from other domains and on modern usage of the various languages. The effect of the lexicon size on the performance of the classifier could also be investigated. It would be interesting to estimate the performance ceiling on LID of short sentences.

Stemming of the lexicon could possibly ensure that the LID generalises better to unseen words. However, stemming in many of the South African languages hasn't been addressed yet.

It should also be interesting to train an end-to-end Deep RNN or CNN to do language ident in the South African context as opposed to manually engineering the two stage classifier.




\bibliographystyle{IEEEtran}
\bibliography{references}


\end{document}